\documentclass{article}

\usepackage{PRIMEarxiv}

\usepackage[utf8]{inputenc} 
\usepackage[T1]{fontenc}    
\usepackage{hyperref}       
\usepackage{url}            
\usepackage{booktabs}       
\usepackage{amsfonts}       
\usepackage{nicefrac}       
\usepackage{microtype}      
\usepackage{lipsum}
\usepackage{fancyhdr}       
\usepackage{graphicx}       
\graphicspath{{media/}}     

\usepackage{tikz}

\usepackage{adjustbox}
\usepackage{multirow}
\usepackage{soul}
\usepackage{subcaption}

\newcommand{\myparagraph}[1]{\paragraph{#1}\mbox{}\\}

\title{Synthetic Context Generation for Question Generation
}

\author{
  Naiming Liu \\
  Rice University \\
  \texttt{nl35@rice.edu} \\
   \And
  Zichao Wang \\
  Adobe \\
  \texttt{jackwa@adobe.com} \\
  \And
  Richard Baraniuk \\
  Rice University \\
  \texttt{richb@rice.edu} \\
}

\begin{document}
\maketitle

\begin{abstract}
Despite rapid advancements in large language models (LLMs), QG remains a challenging problem due to its complicated process, open-ended nature, and the diverse settings in which question generation occurs. 
A common approach to address these challenges involves fine-tuning smaller, custom models using datasets containing background context, question, and answer. 
However, obtaining suitable domain-specific datasets with appropriate context is often more difficult than acquiring question-answer pairs. 
In this paper, we investigate training QG models using synthetic contexts generated by LLMs from readily available question-answer pairs.
We conduct a comprehensive study to answer critical research questions related to the performance of models trained on synthetic contexts and their potential impact on QG research and applications. 
Our empirical results reveal: 1) contexts are essential for QG tasks, even if they are synthetic; 2) fine-tuning smaller language models has the capability of achieving better performances as compared to prompting larger language models; and 3) synthetic context and real context could achieve comparable performances. 
These findings highlight the effectiveness of synthetic contexts in QG and paves the way for future advancements in the field.
\end{abstract}


\section{Introduction}
Recent years witness an increasing presence of Automatic question generation (QG) in various natural language processing applications.
For example, many proposed techniques for large language models (LLMs) leverage the idea of QG to unlock LLMs' reasoning capabilities and mitigate hallucination~\cite{wei2022chain,wang-etal-2022-iteratively}. In  personalized education, QG has the potential to enable custom learning experiences in subjects such as reading comprehension~~\cite{Wolfe1976,Kokku2018,Zhang2022,2206.04187} and reduce the costs and lengths of standard assessment tests~\cite{burstein2021theoretical}. 

Despite the rapid progress in language technologies such as LLMs, QG still remains a challenging problem. This is due to several factors: 1) LLMs are mostly designed to answer questions instead of asking them~\cite{wei2022chain}, making it challenging to design appropriate prompt for QG; 2) question generation is mostly an open-ended process, requiring creativity and deep domain knowledge~\cite{su-etal-2020-multi,liu-etal-2021-mathematical,elsahar-etal-2018-zero}; 3) question generation involves diverse settings, each with its own challenges~\cite{qiu-etal-2020-automatic,liang-etal-2018-distractor}. Because of these reasons, QG remains an active area of research. 

A popular alternative to using LLMs for QG is to fine-tune a smaller, custom model, which can often achieve superior performance on specific tasks than LLMs~\cite{2306.11644}.
However, a challenge in fine-tuning models for QG is the availability of training data, usually in the forms of \{question, answer, context\} triplets~\cite{Wang2018}. Whereas question-answer pairs are relatively easy to obtain, the background text (context) associated with the question-answer pair are not. 
Some existing datasets contain all three elements (contexts, questions, and answers)~\cite{rajpurkar-etal-2018-know}, but a model trained on such dataset often do not appropriately adapt to a specific QG domain such as math word problems~\cite{wang-etal-2021-math} or fairytales~\cite{xu-etal-2022-fantastic}. Obtaining domain-specific dataset is still critical but the contexts are often either difficult to find or unavailable (e.g., copyrighted or behind a paywall).

\vspace{-3pt}
\subsection{Contributions}
\vspace{-3pt}
In this paper, we study the problem of training a QG model from {\it synthetic} contexts, assuming data in the form of only question and answer pairs, without contexts, are readily available. The idea is to generate synthetic contexts, using LLMs, from the question-answer pairs, and then use the generated contexts, together with the question-answer pairs, as training data to train a QG model. 

Our idea is motivated by two recent successes and research trends. First, studies have demonstrated that today's LLMs can already generate highly diverse, creative, and human-like content~\cite{2203.02155,2303.08774}. We posit that they can also generate plausible background texts similar to the real ones. Second, recent research demonstrates the feasibility of training high-performing, albeit smaller, models with synthetic data generated by LLMs~\cite{2212.10560,alpaca,vicuna2023}. These results make it appealing to use LLMs to synthesize contexts and use them for training QG models. 

We conduct a scientific investigation into the feasibility of generating synthetic contexts constitute for traing QG models. Our empirical findings demonstrate the following: 1) QG tasks heavily rely on contexts, whether they are synthetic or not; 2) Fine-tuning smaller language models can lead to superior performance compared to larger language models when prompted; and 3) Synthetic contexts and real contexts can achieve similar levels of performance. These results emphasize the efficacy of synthetic contexts in QG and open avenues for future advancements in this field.

\begin{figure*}[t!]
  \centering
    \includegraphics[width=0.9\linewidth]{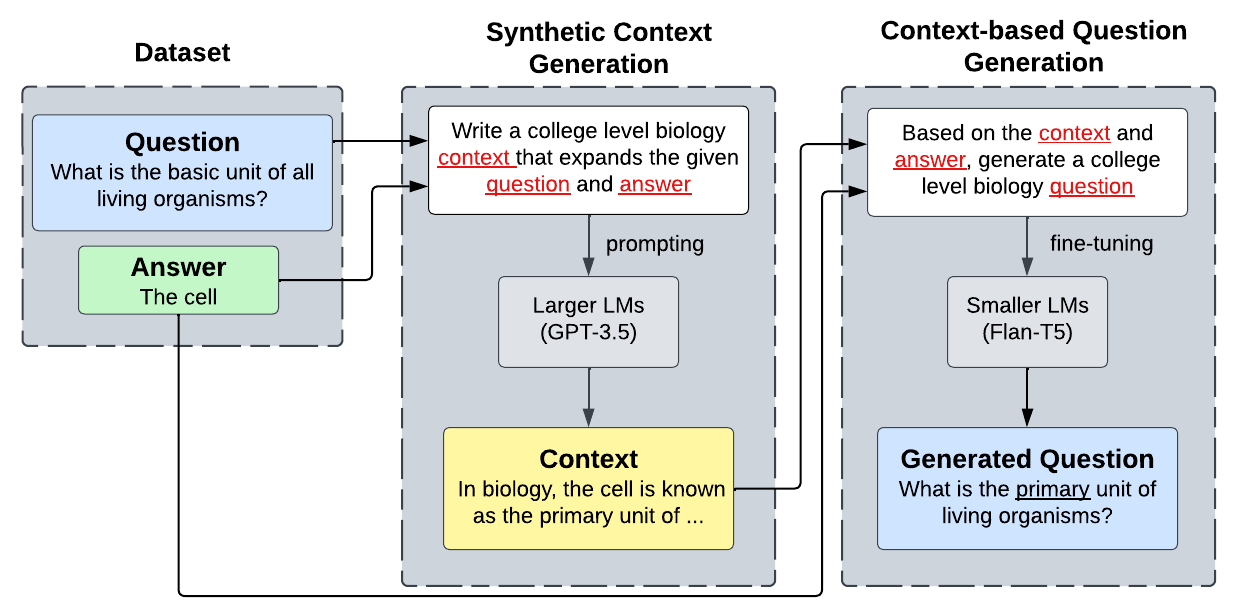}
    \caption{Detailed Overview of Context Generation for Question Generation. We first prompt LLMs to generate synthetic context, then use the generated context and answer to fine-tune smaller LMs for question generation.}
    \vspace{-7pt}
  \label{fig:qg}
\end{figure*}
\section{Approach}
\vspace{-5pt}
Our Approach, context generation for question generation, leverages existing LLMs for creating synthetic context for QG. 
Our method comprises two components: (1) {\bf Synthetic Context Generation}, in which we employ LLMs to produce contextually relevant samples based on given question-answer pairs, and (2) {\bf Context-Based Question Generation}, which uses the synthetic context generated from the previous step, along with real answers, to carry out question generation tasks. A diagram of can be found in Figure~\ref{fig:qg}. 

\vspace{-3pt}
\subsection{Synthetic Context Generation}
\vspace{-3pt}
Consider a set of question-answer pairs $\{(q_1, a_1),  \cdots, (q_n, a_n)\}$, We first prompt an LLM to generate relevant context $c_i$ that could help to answer the question $q_i$ with the given answer $a_i$. We formulate our prompt as 

{\small \texttt{Your job is to write a \{style\} paragraph that significantly expands the given question \{q\_i\} and answer \{a\_i\}.} }

where {\small \texttt{\{style\}}} denotes the particular style of context required to be generated (for instance, for SQuAD dataset, it would be ``wikipedia-style''). Similarly, {\small \texttt{\{q\_i\}}} and {\small \texttt{\{a\_i\}}} are  placeholders that will be replaced with questions and answers respectively. For each $(q_i, a_i)$ pair, the LLM outputs a context $c_i$ that contains relevant background information to support answering the question.

\subsection{Context-based Question Generation}
We adopt a language model (LM) fine-tuning approach for our context-based question generation. After obtaining the context from the previous step, we construct a dataset of triplets $\{(q_1, c_1, a_1),  \cdots, (q_n, c_n, a_n)\}$. Our goal is to generate the ground truth question $q_i$ conditioned on the input context-answer pair $(c_i, a_i)$ using the negative log-likelihood objective function. The input to LM uses the following template: {\small \texttt{Based on the context $c_i$ and answer $a_i$, generate a \{style\} question}}. Training follows the standard causal language modeling objective where the model takes the context and answer as input and aims to predict the question.

\begin{table}[t!]
\centering
\begin{tabular}{ccccc}
\toprule
\textbf{Model} & \textbf{Bleu-4} & \textbf{Meteor} & \textbf{Rouge-l} & \textbf{Bleurt} \\ \midrule
\multirow{2}{*}{\textbf{\begin{tabular}[c]{@{}c@{}}Flan-T5-large\\ (w/o synthetic context)\end{tabular}}} & \multirow{2}{*}{0.035} & \multirow{2}{*}{0.106} & \multirow{2}{*}{0.136} & \multirow{2}{*}{0.227} \\
 &  &  &  &  \\ \midrule
\textbf{Flan-T5-large} & \textbf{0.191} & \textbf{0.397} & \textbf{0.383} & \textbf{0.528} \\
\textbf{davinci-003 (zero)} & 0.079 & 0.262 & 0.212 & 0.419 \\
\textbf{davinci-003 (few)} & 0.116 & 0.309 & 0.269 & 0.507 \\
\textbf{GPT-3.5 (zero)} & 0.107 & 0.352 & 0.256 & 0.488 \\
\textbf{GPT-3.5 (few)} & 0.147 & 0.370 & 0.306 & 0.495 \\ \bottomrule
\end{tabular}
\vspace{3mm}
\caption{Performance of question generation using the OS-bio dataset, where first row denotes QG without context and the rest are QG with synthetic context. Noticeably, fine-tuning smaller language model outperforms prompting larger language models.}
\label{tab:rq1rq2}
\vspace{-3mm}
\end{table}

\section{Experiments}

\paragraph{Dataset} We conducted experiments on two datasets, OpenStax Biology-2e textbook (OS-Bio) \cite{clark2018biology} and SQuAD, the Stanford Question Answering Dataset \cite{rajpurkar2016squad}. OS-bio dataset contains the review questions at the end of each chapter of an introductory-college level biology textbook published by OpenStax, which we curate by ourselves and will be open-sourced.
All results are reported on the test split. 

\paragraph{Experimental Setup}

For synthetic context generation, we use the GPT-3.5 model from OpenAI's API \cite{ouyang2022training} with nucleus sampling \cite{holtzman2019curious}
( $p=1$) and temperature of 0.9. 
We adopt both zero-shot and few-shot in-context learning strategies where we use two pre-selected examples to guide LLMs for generation.
For question generation, we fine-tune pre-trained Flan-T5-large model~\cite{chung2022scaling} for 10 epochs with early stopping. During training, we use synthetic context; during testing, we use real context (if available) in order to examine the QG model's generalizability  and performance in real-world scenarios.
Additionally, for QG on the SQuAD dataset, we randomly select 1000, 5000, 10000 datapoints to generate synthetic context as our training data and then evaluate on the whole test set with real context.\footnote{Due to the costs of OpenAI API, we choose to generate and experiment with synthetic context for a fraction of the training set.} 
More experiment details can be found in Appendix~\ref{app:exp}.

\paragraph{Evaluation} We choose four evaluation metrics including \textbf{BLEU-4}, \textbf{METEOR}, \textbf{ROUGE-L}, and \textbf{BLEURT}, all of which have been widely used in existing QG works.

\begin{table}[t!]
\centering
\begin{tabular}{ccccc}
\toprule
\textbf{Context Type} & \textbf{Bleu-4} & \textbf{Meteor} & \textbf{Rouge-l} & \textbf{Bleurt} \\ \midrule			
\textbf{real} & 0.132 & 0.337 & \textbf{0.356} & 0.457 \\
\textbf{synthetic (zero)} & 0.143 & 0.312 & 0.333 & 0.433 \\
\textbf{synthetic (few)} & 0.151 & 0.324 & 0.347 & 0.443 \\ \midrule
\textbf{real (all)} & \textbf{0.155} & \textbf{0.338} & 0.352 & \textbf{0.459} \\
\bottomrule
\end{tabular}
\vspace{3mm}
\caption{QG performance on the SQuAD dataset trained with real vs. synthetic context generated with zero and few-shot prompting. 
The first three rows takes 1000 samples as the training set, while the last row uses all (87599) datapoints, serving as an uppper bound. 
Remarkably, 1000 synthetic contexts can already yield comparable QG performances compared to all real context as the training set.}
\vspace{-3mm}
\label{tab:rq3}
\end{table}

\subsection{Research Questions and Findings}
We investigate six research questions (RQ) on the use of synthetic context for question generation, and subsequently provide an in-depth analysis based on our experimental results.

\begin{figure}[t!]
  \centering
  \begin{subfigure}[b]{0.49\linewidth}
    \includegraphics[width=\linewidth]{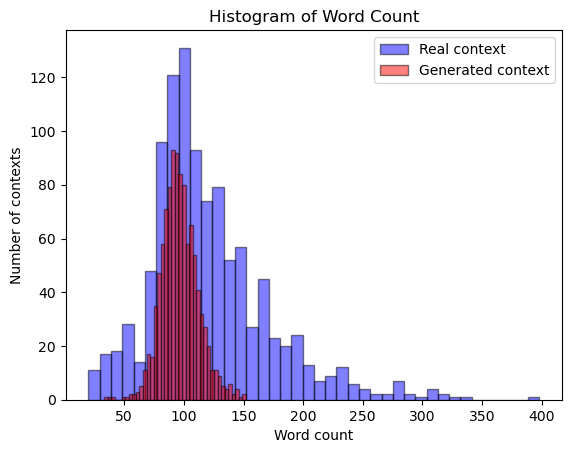}
  \end{subfigure}
  \begin{subfigure}[b]{0.49\linewidth}
    \includegraphics[width=\linewidth]{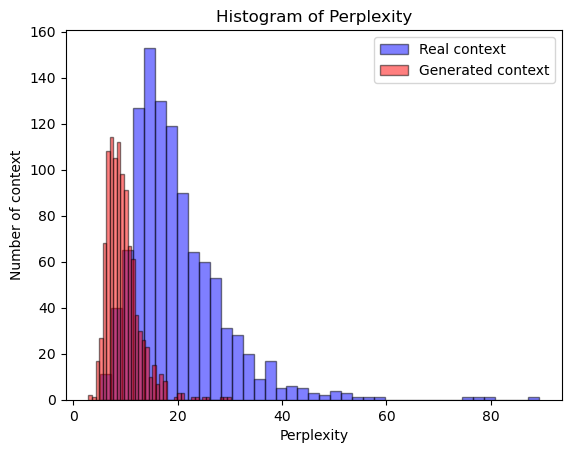}
  \end{subfigure}
  \caption{Word count and perplexity distribution for real and synthetic context generated with few-shot learning.}
  \label{fig:rq4}
\end{figure}

\begin{table*}[t!]
\begin{minipage}{.5\textwidth}
\centering
\begin{tabular}{ccccc}
\toprule
\textbf{Model Type} & \textbf{BLEU-4} & \textbf{Meteor} & \textbf{Rouge-l} & \textbf{Bleurt} \\ \midrule
\textbf{small (S)} & 0.106 & 0.250 & 0.282 & 0.371 \\
\textbf{medium (S)} & 0.124	& 0.296 &0.325 & 0.426 \\
\textbf{large (S)} & 0.132 & 0.324 & 0.347 & 0.453 \\ \midrule
\textbf{small (R)} & 0.100 & 0.241 & 0.267 & 0.359 \\
\textbf{medium (R)} & 0.131	& 0.291	& 0.317	& 0.414 \\
\textbf{large (R)} & 0.151 & 0.324 & 0.347 & 0.443 \\ 
\bottomrule
\end{tabular}
\vspace{2mm}
\caption{Performance of QG on SQuAD for different model size. The small, medium and large represents Flan-T5-small, -medium and -large respectively. (S) means synthetic context generated using few-shot learning, while (R) denotes real context.}
\label{tab:rq5-1}
\end{minipage}
\hspace{3mm}
\begin{minipage}{.5\textwidth}
\centering
\begin{tabular}{ccccc}
\toprule
\textbf{Data Type} & \textbf{BLEU-4} & \textbf{Meteor} & \textbf{Rouge-l} & \textbf{Bleurt} \\ \midrule
\textbf{small (S)} & 0.151 & 0.324 & 0.347 & 0.443 \\
\textbf{medium (S)} & 0.135 & 0.308	& 0.319	& 0.420 \\
\textbf{large (S)} & 0.109 & 0.295 & 0.305 & 0.407 \\ \midrule
\textbf{small (R)} & 0.132 & 0.324 & 0.347 & 0.453 \\
\textbf{medium (R)} & 0.157 & 0.337	& 0.356	& 0.457 \\
\textbf{large (R)} & 0.149 & 0.328 & 0.351 & 0.456 \\
\bottomrule
\end{tabular}
\vspace{2mm}
\caption{ Performance of QG on SQuAD for different training set size. The small, medium and large represents 1000, 5000, 10000 datapoints respectively. (S) means synthetic context generated using few-shot learning, while (R) denotes real context.}
\label{tab:rq5-2}
\end{minipage}
\vspace{-3mm}
\end{table*}

\myparagraph{RQ1: Is context necessary for question generation, even if the context
is synthetic?}
As demonstrated in Table~\ref{tab:rq1rq2}, the performance of QG without context is significantly worse than QG with synthetic context for fine-tuning Flan-T5-large model. This outcome can be attributed to the fact that the presence of context, regardless of its authenticity, offers vital information for generating meaningful and appropriate questions. Hence, the inclusion of context proves to be crucial for QG task, even if the provided context are synthetic. 

\myparagraph{RQ2: How does fine-tuned small language models compare to prompting LLMs on QG?}
We also utilize OS-bio dataset to investigate the performance differences between fine-tuning smaller language models and prompting larger language models. We employ Flan-T5-large model with 780M parameters for fine-tuning, while adopt GPT-based models of 175B parameters for zero-shot and few-shot prompting. As illustrated in Table~\ref{tab:rq1rq2}, when synthetic context is available, fine-tuning smaller language models could achieve better performance compared to prompting larger language models. Also as anticipated, when we use few-shot learning, the model performance improves as compared to zero-shot learning, especially when the context is present. These comparisons demonstrate the superior performance of fine-tuned small models compared to LLMs on the QG task.

\myparagraph{RQ3: How does the question generated from real context compared to fake context?} 
We examine the QG performance difference between models trained on real and synthetic contexts. We compare three types of contexts: real context from SQuAD dataset, synthetic context generated using GPT-3.5 API with zero-shot and few-shot learning. Our results, as shown in Table~\ref{tab:rq3}, demonstrate that models trained on merely 1000 synthetic contexts can yield strikingly comparable QG performances compared to models trained on all real context. 
In RQ5, we further analyze the effect on the number of synthetic training examples.

\myparagraph{RQ4: What is the synthetic contexts' quality? } 
We perform two assessments to examine the quality of the generated contexts.
First, we assess the disparities in word count and perplexity (using GPT-2 model \cite{radford2019language}), with the results presented in Figure~\ref{fig:rq4}. Despite the distribution of word count between real and synthetic context being relatively similar, it is worth noting that the synthetic context exhibits lower perplexity in general as compared to real context, suggesting that the synthetic contexts align better with the language model's distribution and are less diverse than real contexts.

Second, we conduct a question answering (QA) evaluation task to explore whether synthetic context contains the useful information to answer the given question. We find that that 84\% of the synthetic contexts contain the answer phrase. Using a QA model\footnote{We use the Roberta-base: \url{https://huggingface.co/deepset/roberta-base-squad2}}, we also find that synthetic context enables the QA model to answer question correctly 61\% of the time with 0.77 F1 score. 
We further manually examined 100 instances of synthetic context that did not obtain an exact match, and upon inspecting these, we found that only seven contexts contained confusing information for question-answering. 
These results indicate a reasonable level of information inclusion of the synthetic context, suggesting its effectiveness for question generation.


\myparagraph{RQ5: What is the impact of model size and training data size for question generation?} 
Table~\ref{tab:rq5-1} evidently shows that an increase in model size leads to improved QG performance, regardless of the context type. However, Table~\ref{tab:rq5-2} shows that the effect of training dataset size on performance varies on the nature of the context. For real context, an increase in the training dataset tends to results in improved QG performance. 
However, we observe a decline in QG performance when the training dataset size increases for the synthetic dataset. 
We hypothesize that more synthetic data introduce more noise and inconsistencies into the learning process, leading to an increased mismatch between the training (synthetic contexts) and test (real contexts) data.
A potential mitigation could be few-shot prompting for context generation, which achieves the second strongest result in Table~\ref{tab:rq1rq2} and could potentially reverse the diminishing trends in QG as the synthetic training dataset size increases. 
\begin{figure}[t!]
  \centering
  \begin{subfigure}[b]{0.49\linewidth}
    \includegraphics[width=\linewidth]{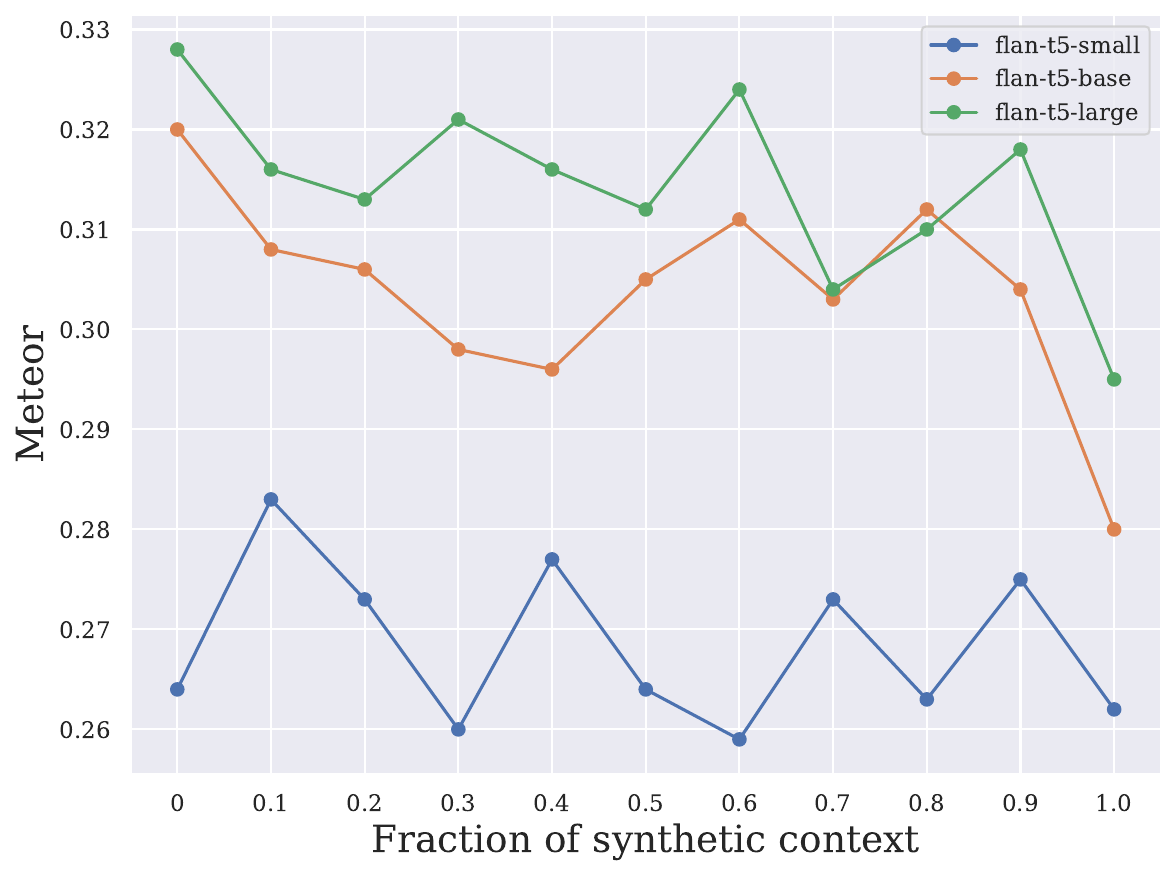}
  \end{subfigure}
  \begin{subfigure}[b]{0.49\linewidth}
    \includegraphics[width=\linewidth]{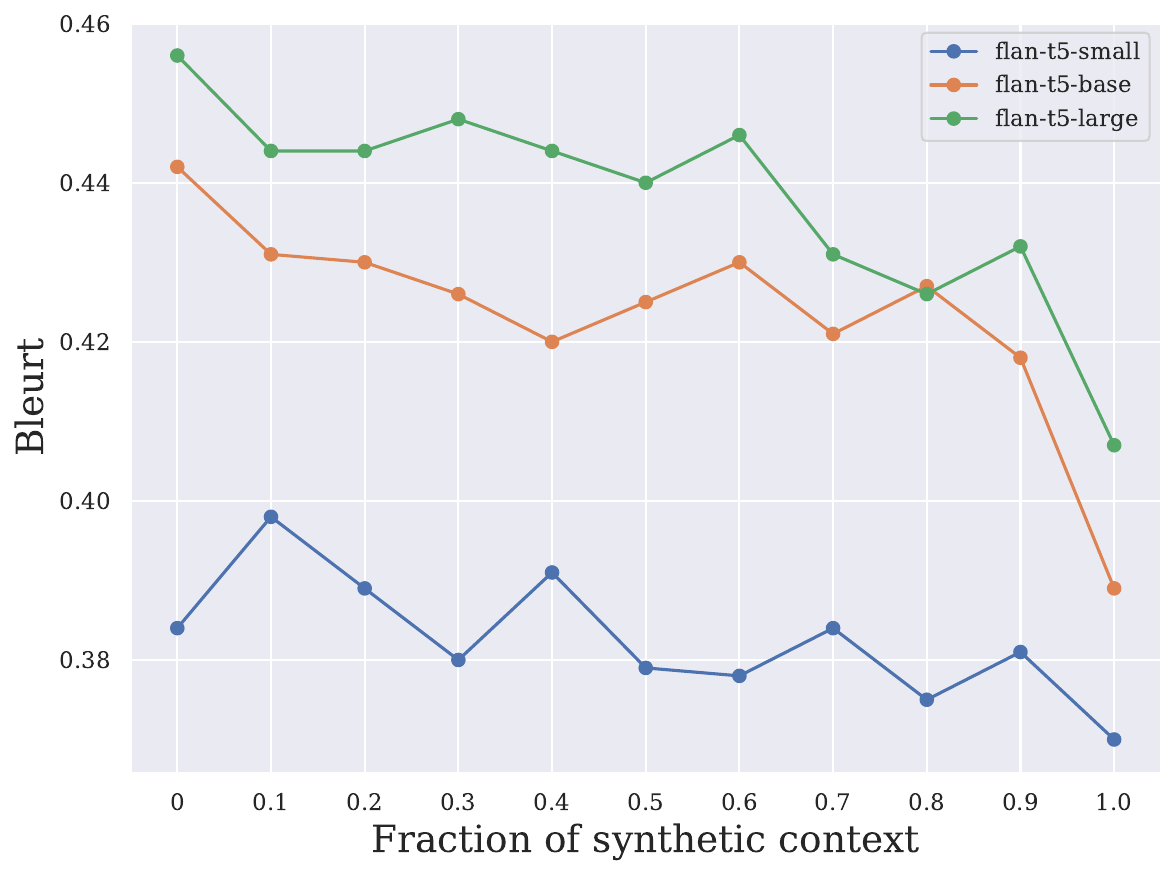}
  \end{subfigure}
  \caption{Performance of QG as the fraction of synthetic context increases. We present Meteor and Bleurt evaluation metric here.}
  \label{fig:rq6}
  \vspace{-5mm}
\end{figure}

\label{sec:rq6}
\myparagraph{RQ6: What happens when the real and synthetic context are mixed?}
In order to further explore the effects of synthetic context for QG, we conducted an experiment with an interpolation between real and synthetic context generated with few-shot prompting on the subset of 10,000 SQuAD datapoints. We aim to investigate the changes in QG performance as the proportion of synthetic context increases within the training dataset. The results in Figure~\ref{fig:rq6} generally reveal a generally decreasing trends in performance, as anticipated, when the proportion of synthetic context increases. Another intriguing observation is that when the training dataset contains only synthetic context, the performance experiences a substantial decline, which implies the effectiveness of incorporating some real context within the training dataset. Even the inclusion of a minimal portion (approximately 10\%) of real context mixed with a predominantly synthetic dataset appears to boost the QG performance. A possible explanation is that integrating a small yet substantial portion of real context could help with reducing the gap between training and testing context. Further research could explore the ideal proportions of real and synthetic context to not only ensure model performance but also mitigate the efforts of obtaining real context.

\section{Related Work}

Automatic question generation (QG) is a complex task that potentially 1) involves the generation of many different types of questions including factual~\cite{Wang2018}, multi-hop~\cite{su-etal-2020-multi}, multiple-choice~\cite{qiu-etal-2020-automatic}, and multi-document~\cite{Cho2021}; 2) spans multiple domains such as literacy~\cite{xu-etal-2022-fantastic}, math~\cite{wang-etal-2021-math}, science~\cite{Welbl2017}, language learning~\cite{burstein2021theoretical}, and customer services~\cite{1811.10686}; 3) requires controllability in terms of difficulty~\cite{Cheng2021} or multi-modality~\cite{lu2022learn}. Furthermore, QG is generally considered a more challenging task than question answering, where the former usually requires more creativity, intentionally, and art~\cite{Schein2021-nz,Le_Baron_Payne2014-xy}. Because of the aforementioned complexity and challenges, currently large language models' performance in QG remains under-explored and unsatisfactory, and their utility is commonly limited in simple factual QG or as an auxiliary task, rather than the objective, to facilitate another task such as QA. Fine-tuning a smaller, QG-specific model is promising approach to improve domain-specific performance, as a few recent works have demonstrated. Our present work fits in the fine-tuning paradigm and investigates the feasibility of synthesizing the contexts as part of the QG fine-tuning data because the contexts are important to QG but can be challenging to obtain. 

Our work is also related to the recent lines of work demonstrating the promise to use LLM generated synthetic data for fine-tuning smaller models in situations such as instruction tuning~\cite{2212.10560,alpaca,vicuna2023}. Some works also show, empirically, that synthetic data leads to stable and useful fine-tuning as long as the generated data is close to the real data~\cite{2209.03942}. These prior works inspire our present work and motivate us to adapt the synthetic data fine-tuning paradigm to the niche direction of QG. Methodology-wise, our work shares many merits with prior works. However, we investigate a few questions that are unique to QG, such as the necessity of contexts during fine-tuning and the quality of the synthesized contexts.

Our investigations reveals insights and caveats of synthesizing training data specifically for QG, offering practical suggestions and opening further research opportunities to study the various approaches to fine-tune language models for QG.

\section{Conclusion}

In this paper, we investigate the effects of using synthetic context generated by LLMs to train QG models. We provide in-depth analysis to address critical research questions associated with this approach. Our experimental results demonstrate that the importance of synthetic context for LM-based question generation task as they can achieve comparable performance levels to real context. Furthermore, our findings revealed that, given the presence of related context—irrespective of its origin—fine-tuning smaller language models can yield superior performance compared to prompting larger language models.

\section*{Acknowledgments}
This work was supported by NSF grants 1842378, ONR grant N0014-20-1-2534, AFOSR grant FA9550-22-1-0060, and a Vannevar Bush Faculty Fellowship, ONR grant N00014-18-1-2047.

\bibliographystyle{abbrv}  
\bibliography{reference}  

\appendix
\section{Experimental Details}
\label{app:exp}
\subsection{OS-bio Dataset}
We provide several examples of question-answer pairs in OS-bio dataset, along with their generated context in Table~\ref{tab:os-bio}. 
For the OS-bio dataset, we attempted to manually extract relevant information from the OS-bio textbook to serve as real context. However, upon closer examination, we found that only approximately 50\% of the samples in the test set had related contexts within the textbook. The remaining samples either could only be inferred from the textbook information or were spread across the entire chapter without corresponding to a specific section. Moreover, the QG performance utilizing \textit{extracted} contexts was not satisfactory, which underscores the importance of generating synthetic contexts using LLMs for QG task.

\subsection{Prompt design}
In this section, we provide some additional prompts for context and question generation when using prompting strategy. Initially, for the context generation with OS-bio dataset, the {\small \texttt{\{style\}}} placeholder is replaced by {\small \texttt{an introductory college level scientific paragraph about biology}}, while for SQuAD dataset, {\small \texttt{\{style\}}} is {\small \texttt{wikipedia-style paragraph}}. Additionally for the few-shot learning with SQuAD dataset, we add a special indicator \textbf{title} to ensure the generated context are relevant. The new prompts are changed to:

{ \small\texttt{Your job is to write a wikipedia-style paragraph on a specific topic. Your written paragraph should contains the answer to a question that asks about certain information related to the topic. The user will first provide the topic, question, and answer and some example paragraphs.}}

Secondly, for the prompting-based QG with OS-bio dataset, we use the following prompt, 

{\small \texttt{Based on the context below, generate an introductory college level biology question with \{a\} as the answer.}}

\subsection{Additional Experimental Setup}
Regarding synthetic context generation with the SQuAD dataset, we employ both zero-shot and few-shot in-context learning strategies, utilizing two pre-selected examples to guide LLMs. A few-shot learning example is shown below:
\vspace{2pt}

{\small \texttt{``title": ``Solar\_energy",}

\texttt{``context": ``Solar power is the conversion of sunlight into electricity, either directly using photovoltaics (PV), or indirectly using concentrated solar power (CSP). CSP systems use lenses or mirrors and tracking systems to focus a large area of sunlight into a small beam. PV converts light into electric current using the photoelectric effect.",}

\texttt{``question": ``What method does the photovoltaics system use to turn light into electricity?",}

\texttt{``answer": ``photoelectric effect"}
}

For the question generation task, we conduct all training on a single NVIDIA Quadro RTX 8000 GPU, applying a consistent training setup across all QG tasks. Specifically, we adopt a learning rate of 0.0003 and train the model for 10 epochs and implement early stopping if the validation loss fails to improve over the most recent 3 epochs. Additionally, we use a batch size of 8. All these training procedures are standard in fine-tuning question generation models.

\begin{table*}[t!]
\centering
\small
\begin{tabular}{p{0.2\linewidth}p{0.2\linewidth}p{0.6\linewidth}}
\toprule
\textbf{Question} & \textbf{Answer} & \textbf{Synthetic Context} \\ \midrule
Bacteria that lack fimbriae are less likely to \_\_\_\_. & adhere to cell surfaces & Bacteria possess a variety of extracellular appendages that allow them to attach to surfaces, including fimbriae. \hl{Fimbriae are thin, hair-like projections composed of proteins that allow bacteria to adhere to surfaces} such as cell membranes in the body. Bacteria lacking fimbriae are less likely to adhere to cell surfaces and therefore less likely to cause infection. This is due to the fact that fimbriae act as receptors that allow bacteria to recognize and bind to certain molecules on the surface of a cell. Without these receptors, bacteria lack the ability to attach to cells and therefore cannot colonize or cause infection. Thus, a lack of fimbriae can make bacteria less likely to cause infection. \\ \midrule
Meiosis usually produces \_\_\_\_ daughter cells. & four haploid & Meiosis is an important process of cell division that is vital for the production of gametes in sexually-reproducing species. \hl{It is a unique type of cell division as it results in the production of four haploid daughter cells}, each with only one set of chromosomes. This is in contrast to mitosis, which produces two daughter cells with the same genetic information as the parent cell. During meiosis, the chromosomes within the parent cell pair up and exchange genetic material with their partner, a process known as crossing over. This leads to an increase in genetic diversity which is beneficial for the survival of the species. \\ \midrule
Population genetics is the study of: & how selective forces change the allele frequencies in a population over time & \hl{Population genetics is a field of study that investigates how selective forces, such as natural selection and genetic drift, affect the allele frequencies of a population over time}. By examining the changes in allele frequencies of a population over multiple generations, population geneticists can gain insight into the evolutionary history of the population, such as which selective forces have been most influential in shaping its genotype. Additionally, population genetics can provide clues about the current health of the population and how it may be affected by environmental change. \\
\bottomrule
\end{tabular}
\caption{Examples of question-answer pairs in OS-bio dataset, together with their synthetic context. We highlight the portions that include the provided answers.}
\label{tab:os-bio}
\end{table*}

\end{document}